\documentclass[lettersize,journal]{IEEEtran}

\usepackage{cite}
\usepackage{amsmath,amssymb,amsfonts,amsthm}
\usepackage{graphicx}
\usepackage{textcomp}
\usepackage{xcolor}
\usepackage{tikz}
\usepackage{bm}
\usepackage{xurl} 
\usepackage{hyperref}
\usepackage{booktabs}
\usepackage{multirow}
\usepackage{lipsum}
\usepackage[linesnumbered, ruled, noline]{algorithm2e}

\ifCLASSOPTIONcompsoc 
\usepackage[caption=false,font=normalsize,labelfont=sf,textfont=sf]{subfig}
\else
\usepackage[caption=false,font=footnotesize]{subfig}
\fi
\newtheorem{definition}{Definition}
\AtBeginDocument{%
  \providecommand\BibTeX{{%
    \normalfont B\kern-0.5em{\scshape i\kern-0.25em b}\kern-0.8em\TeX
 }}%
}

\begin{document}

\title{Towards Data-driven Nitrogen Estimation in Wheat Fields using Multispectral Images}

\author{Andreas Tritsarolis, Tomaž Bokan, Matej Brumen, Domen Mongus, Yannis Theodoridis%
\thanks{A. Tritsarolis (\url{andrewt@unipi.gr}) and Y. Theodoridis (\url{ytheod@unipi.gr}) are with the Department of Informatics, University of Piraeus, Piraeus, Greece}%
\thanks{T. Bokan (\url{tomaz.bokan@itc-cluster.com}) is with the Innovation Technology Cluster, Murska Sobota, Slovenia}%
\thanks{M. Brumen (\url{matej.brumen@um.si}) and D. Mongus (\url{domen.mongus@um.si}) is with the Department of Electrical Enginnering \& Computer Science, University of Maribor, Maribor, Slovenia}%
}

\maketitle

\begin{abstract}
    The modernization of agriculture has motivated the development of advanced analytics and decision-support systems to improve resource utilization and reduce environmental impacts. Targeted Spraying and Fertilization (TSF) is a critical operation that enables farmers to apply inputs more precisely, optimizing resource use and promoting environmental sustainability. However, accurate TSF is a challenging problem, due to external factors such as crop type, fertilization phase, soil conditions, and weather dynamics. In this paper, we present TerrAI, a Neural Network-based solution for TSF, which considers the spatio-temporal variability across different parcels. Our experimental study over a real-world remote sensing dataset validates the soundness of TerrAI on data-driven agricultural practices.
\end{abstract}


\section{Introduction}\label{sec:Introduction}
    Agriculture has witnessed rapid modernization in recent decades, propelled by technological innovations aimed at enhancing productivity, reducing costs, and promoting environmental sustainability. Targeted Spraying and Fertilization (TSF) has emerged as a critical operation that optimizes resource allocation and minimizes environmental impacts \cite{MULLA2013358}.  
    In particular, TSF aims to estimate the Nitrogen (N) requirements, taking into account land characteristics (e.g., soil state, weather forecasts) of a given parcel to reduce fertilizer use and promote sustainability.

    However, accurate TSF is a challenging task influenced by factors such as crop type, growth stage, as well as soil and weather conditions \cite{6844831}. Traditional TSF approaches often rely on manual observations or rudimentary sensor data, which are limited in both scale and accuracy \cite{s18082674}. These limitations have prompted researchers to explore data-driven solutions that leverage advances in machine learning and computer vision. Machine Learning (ML) approaches, Convolutional Neural Networks (CNNs) in particular, have demonstrated considerable potential in extracting spatial and contextual information from agricultural data, enabling more efficient decision-making \cite{KAMILARIS201870}. In this context, we propose TerrAI, an efficient CNN-based solution for TSF. Our experimental study on a real-world remote sensing dataset, validates the effectiveness of our approach on improving resource utilization, and by extension, contributing to environmental sustainability. In summary, the main contributions of this paper are as follows: 

    \begin{itemize}
        \item We model TSF as a spatially-aware ML task.
        \item We introduce TerrAI, an efficient CNN-based framework for TSF.
        \item We demonstrate the soundness of the proposed architecture, in terms of accuracy, using a real-world remote sensing dataset.
    \end{itemize}
    %
    The rest of this paper is organized as follows: Section \ref{sec:RelatedWork} discusses related work. Section \ref{sec:Background_and_Definitions} formulates the problem at hand and describes the proposed TerrAI framework\footnote{For reproducibility purposes, the entire experimental protocol (data preprocessing, model training, inference, and evaluation pipelines) is available at \url{https://github.com/DataStories-UniPi/TerrAI}.}. Section \ref{sec:experiments} presents the results of our experimental study, where we demonstrate the efficacy of our solution. Finally, Section \ref{sec:conclusion} concludes the paper, also giving hints for future work.

\section{Related Work}\label{sec:RelatedWork}    
    \begin{figure*}[!ht]
        \centering
        \includegraphics[width=\textwidth]{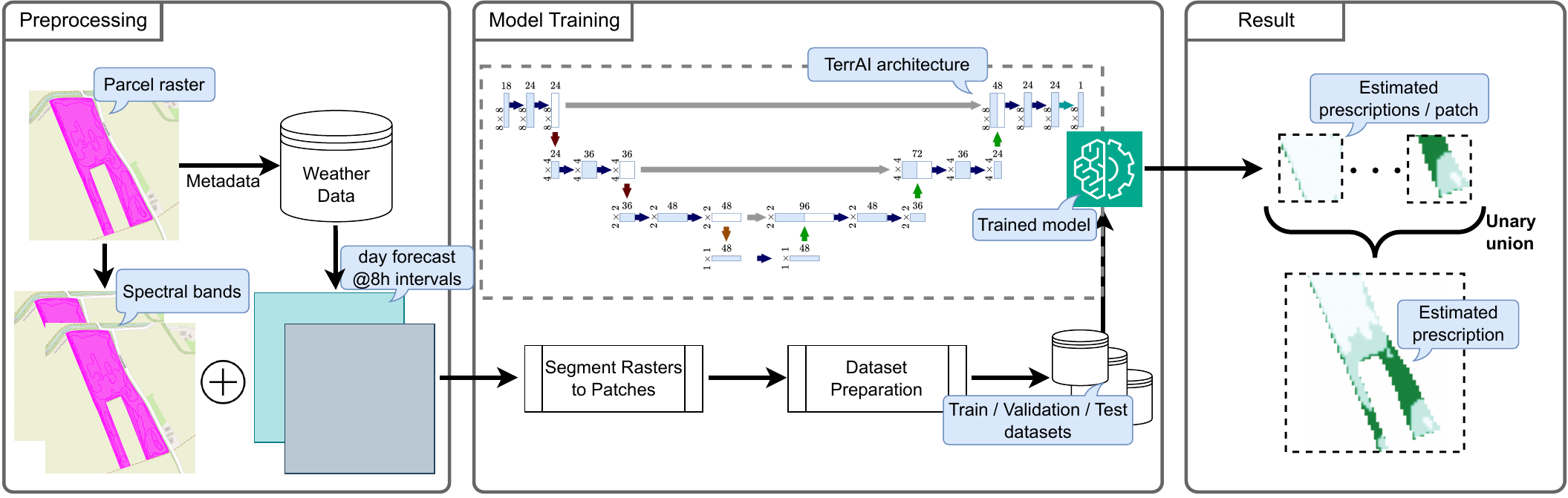}
        \caption{Architecture overview of the proposed TerrAI framework.}
        \label{fig:TerrAI-Workflow}
    \end{figure*}
    
    Considering the agriculture domain, current state-of-the-art includes an adequate number of research works \cite{10.1007/978-981-15-7106-0_51,DBLP:journals/air/LeiYYSWF24,KATTENBORN202124}.
    Regarding ML-based methods for fertilizer estimation, Yamashita et al. \cite{Yamashita2020} evaluate the performance of various regression models for predicting Nitrogen and Chlorophyll content in tea leaves using hyperspectral reflectance data. Cerasola et al. \cite{CERASOLA2025100802} experiment with three regression algorithms to estimate the optimal nitrogen quantity to be supplied in fertigated processing tomatoes through hyperspectral imaging.
    Towards more recent works, Gao et al. \cite{rs15205013} focus on developing a Nitrogen content prediction model using hyperspectral data and similar regression algorithms for the accurate fertilization of citrus trees.
    
    Moving to CNN-based encoder-decoder based architectures, recent works include, but are not limited to, Maurya et al. \cite{MAURYA2023102078} who propose a variant of the well-known U-Net model \cite{DBLP:conf/miccai/RonnebergerFB15}, as well as introduce a novel dataset for the task of semantic segmentation of satellite images into five agricultural land categories. Guo et al. \cite{Guo2025} propose CMTNet, a novel dual-branch network based on CNN and Transformers, in order to enhance crop classification precision and efficiency in complex agricultural environments.
    
    Closer to our work, Sajindra et al. \cite{SAJINDRA2024100395} propose a deep learning model in order to estimate the Nitrogen, Phosphorus, and Potassium content of the soil by analyzing the growing characteristics of cabbage (Brassica oleracea) plants. In terms of CNN-based encoder-decoder architectures, Li et al. \cite{DBLP:journals/cea/LiSMHSHDDL24} propose SCLNC-Net, a novel deep learning framework that combines Attention U-Net \cite{SCHLEMPER2019197} and MobileNetV3 \cite{9008835} to estimate Nitrogen concentration in rice leaves using mobile phone images.
    
    Compared to the aforementioned works, our approach employs a U-Net-based architecture that processes hyperspectral images of Wheat fields to predict their corresponding Nitrogen application rate ($\text{kg} \cdot N / ha$). By framing the problem as a spatially-aware regression task, the model can deliver accurate, parcel-level prescription maps while avoiding the computational overhead and complexity of multitask frameworks.

\section{Problem Formulation and Proposed Methodology}\label{sec:Background_and_Definitions}
    In this section, we formulate the TSF problem and describe the proposed TerrAI solution.

    \subsection{Problem Definition}
        Before we proceed to the actual formulation of the problem, we provide some preliminary definitions.

        \begin{definition}[Soil-health]\label{def:SoilHealth}
            The soil health of a parcel $l$ at fertilization phase $k$ is denoted as a tensor $S_l^k \in \mathbb{R}^{C \times H \times W}$, where $\{S_l^k\}_{ij} \in \mathbb{R}^C, 1 \leq i \leq H, 1 \leq j \leq W$, corresponds to a scaled region of the parcel.
        \end{definition}
        
        \begin{definition}[Prescription map]\label{def:PrescriptionMap}            
            The prescription map of a parcel $l$ at fertilization phase $k$ is denoted as a matrix 
            $P_l^k \in \mathbb{R}^{1 \times H \times W}$, where each entry $\{P_l^k\}_{ij}, 1 \leq i \leq H, 1 \leq j \leq W$ corresponds to the Nitrogen application rate applied to a scaled region of the parcel.
        \end{definition}

        \begin{definition}[Targeted Spraying and Fertilization - TSF]\label{def:TargetedSprayingFertilization}
            Given a dataset $D = \{(S_l^k, P_l^k)\}$ of historical soil-health and prescription map pairs, the goal of the TSF problem is to train a data-driven model over $D$, which will be able to estimate their corresponding prescription map $\hat{P_l^k}$, given their current soil-health state tensor $S_l^k$.
        \end{definition}

    \subsection{Methodology}\label{subsec:Methodology}           
        Figure \ref{fig:TerrAI-Workflow} illustrates the reference architecture of our TerrAI framework. To address the TSF problem, we use the well known U-Net \cite{DBLP:conf/miccai/RonnebergerFB15} architecture for estimating the quantity of fertilizer in a farming area, given its current soil-health status. 
        In brief, a U-Net is a symmetric encoder-decoder convolutional network originally designed for biomedical image segmentation. The encoder progressively downsamples the input while increasing the number of feature channels, and the decoder upsamples the representations, using skip connections to concatenate high-resolution encoder features with corresponding upsampled feature maps. This symmetric design enables precise spatial localization while preserving contextual information, making it well suited for precision agriculture tasks such as the TSF problem.
        
        The input consists of multi-spectral satellite images with 18 channels (i.e., spectral bands): near-infrared (NIR), red (R), green (G), blue (B), and weather condition indices (forecasts) at eight-hour intervals, while the output consists of an image with the same dimensions as the input and one output channel, which corresponds to the amount of fertilizer to be used at each region of the specified parcel.
     
        In order to train TerrAI, we use the Adam \cite{DBLP:journals/corr/KingmaB14} optimization algorithm with learning rate $\eta = 10^{-3}$ and the Root Mean Squared Error (RMSE) loss function for 200 epochs. To prevent over-fitting, we use the well-known early stopping \cite{DBLP:series/lncs/Prechelt12} mechanism with a patience of 10 epochs. In order to avoid calculating loss for the padding of the input images, we restrict RMSE to the areas where we have known observations, or $\{P_l^k\}_{ij} \neq \texttt{no\_data}, 1 \leq i \leq H, 1 \leq j \leq W$. The \texttt{no\_data} flag in hyperspectral images defines areas that are irrelevant to our area-of-interest (e.g., out-of-bounds).

\section{Experimental Study} \label{sec:experiments}
    In this section, we evaluate the efficiency of TerrAI using a real-world remote sensing dataset, and present our experimental results.

    \begin{figure*}[!ht]
        \centering
        \includegraphics[width=\textwidth]{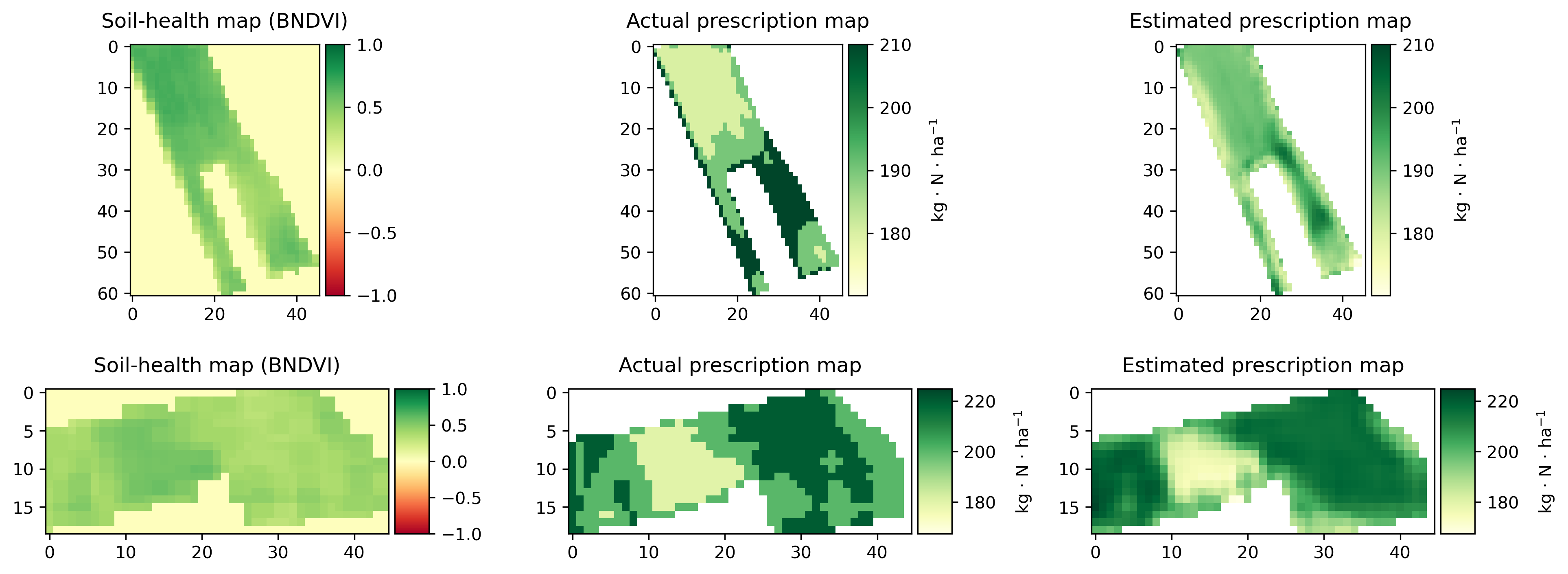}
        \caption{Estimation of the prescription map for two randomly selected test parcels from the ITC dataset using TerrAI. Left: input soil-health tensor (one spectral image); middle and right: actual and estimated prescription maps.}
        \label{fig:TSF-TerrAI-TestSet-Results} 
    \end{figure*}

    \subsection{Experimental Setup, Datasets and Preprocessing}\label{subsec:ExperimentalSetup}
        In our experimental study, we use a real-world remote sensing dataset, referred to as ``ITC''\footnote{The (proprietary) dataset has been kindly provided by ITC and University of Maribor for research purposes, in the context of the EU Horizon project Green.Dat.AI (\url{https://greendatai.eu}).} dataset. Given the specificity of the problem and the particular approach we follow, to the best of our knowledge there are no additional open datasets available for further evaluation.

        The dataset consists of $50$ multi-spectral (NIR, Red, Green, Blue) soil-health images taken from $35$ farming parcels located in the north-eastern part of Slovenia (lon. in $\lbrack 15.85, 16.50 \rbrack$, lat. in $\lbrack 46.51, 46.79 \rbrack$) across three distinct fertilization phases, approximately once every 3 months, between April 7, 2021, to May 8, 2023. The images have a spatial resolution of 10 meters and were captured using a 12-bit multi-spectral instrument (MSI), allowing for light intensity values ranging from 0 to 4095. All experiments were conducted in Python using PyTorch and run on a server with 1 Nvidia A100 GPU, 128 CPUs, and 1TB RAM.
            
        Our data (pre-)processing stage consists of two phases. The first phase (data cleaning) is a process consistently followed in the literature due to the noise, irregularity of image size, etc. that are typical in multi-spectral image datasets. The second phase (data preparation for model training), uses the output of the first phase as its input and is performed for ML-specific purposes, i.e., to feed its output into our TerrAI architecture, as it is presented in Section \ref{subsec:Methodology}. 

        The preprocessing stage involves two main steps: (i) removing outliers based on fertilizer usage using the inter-quartile range (IQR) method, and (ii) performing feature engineering by creating additional spectral bands. To manage varying image sizes, images are segmented into overlapping $8 \times 8$ patches with a $1 \times 1$ stride, representing approximately $80 m^2$ of farmland. The multi-spectral images are further enriched with vegetation indices (e.g., BNDVI \cite{10.1016/S1672-63080760027-4}) and weather forecast data from nearby sensors at 8-hour intervals on the fertilization day. To prevent noise from out-of-bounds values, we set the \texttt{no\_data} flag value from $-9999$, as provided in the dataset, to $0$, utilizing the zero property in multiplication.
    
        To enrich the training dataset and reduce overfitting, data augmentation is applied by randomly flipping image patches horizontally and/or vertically with a 50\% probability. To address differences in units across spectral bands, z-score standardization is used by subtracting the mean and dividing by the variance of each band. The label for each patch is taken from the corresponding section of the prescription map, representing the fertilizer amount per region. These labels go through the same transformations to ensure consistency. Since labels are used solely in the learning objective and not as model input, the value of the \texttt{no\_data} flag remains as-is.

    \subsection{Experimental Results on TerrAI}
    
        \begin{table*}[!ht]
            \centering
            \caption{Comparison of three different TerrAI versions with respect to model size, prediction error, and energy consumption (lower is better).}
            \label{tab:terrai_size_vs_energy_vs_accuracy}
            
            \renewcommand{\arraystretch}{1.3}
            \resizebox{\textwidth}{!}{%
                \begin{tabular}{@{}lllllllllll@{}}
                    \toprule
                    &                                  
                    &                                        
                    \multicolumn{3}{c}{Metrics / Patch} & 
                    \multicolumn{3}{c}{Metrics / Prescription map} &
                    \multicolumn{3}{c}{Green dimension of TerrAI}
                    \\ \cmidrule{3-11}
                    \multirow{2}{*}{Model size} & 
                    \multirow{2}{*}{\#Parameters} & 
                    \multicolumn{1}{c}{\begin{tabular}[c]{@{}c@{}}RMSE \\ ($\text{kg} \cdot \text{N} \cdot \text{ha}^{-1}$)\end{tabular}} & 
                    \multicolumn{1}{c}{MAPE (\%)} & \multicolumn{1}{c}{SMAPE (\%)} & 
                    \multicolumn{1}{c}{\begin{tabular}[c]{@{}c@{}}RMSE \\ ($\text{kg} \cdot \text{N} \cdot \text{ha}^{-1}$)\end{tabular}} & 
                    \multicolumn{1}{c}{MAPE (\%)} & \multicolumn{1}{c}{SMAPE (\%)} &
                    \begin{tabular}[c]{@{}c@{}}Energy consumption \\ per run (J)\end{tabular} &
                    \begin{tabular}[c]{@{}c@{}}Annual savings \\ vs. larger variant (kWh)\end{tabular} &
                    \begin{tabular}[c]{@{}c@{}}CO$_2$ equivalent \\ (g CO$_2$e)\end{tabular}
                    \\ \midrule
                    small & 
                    16,773 & 
                    14.54 & 
                    6.12 & 
                    6.33 & 
                    30.60 & 
                    16.42 & 
                    13.60 &
                    \textbf{16,619} &
                    \textbf{4.60 × 10\textsuperscript{-3}} &
                    \textbf{0.76}
                    \\
                    baseline & 
                    207,421 & 
                    \textbf{12.31} & 
                    \textbf{5.31} & 
                    \textbf{5.23} & 
                    \textbf{21.26} & 
                    \textbf{9.72} & 
                    \textbf{9.16} &
                    33,172 &
                    5.44 × 10\textsuperscript{-3} &
                    0.90
                    \\
                    large & 
                    1,063,981 & 
                    12.73 & 
                    5.35 & 
                    5.36 & 
                    22.25 & 
                    10.84 & 
                    9.94 &
                    52,769 &
                    --- &
                    --- 
                    \\ \bottomrule
                \end{tabular}
            }
        \end{table*}
        
        Following the preprocessing outlined in Section \ref{subsec:ExperimentalSetup} over wheat crops during the second fertilization phase (which constitutes the majority of the data), we acquire 44,953 patches. These are split into training, validation, and test sets using a 60:20:20\% stratified ratio based on the average fertilizer quantity per patch. To prevent data leakage, i.e., patches of images from the test set to be shuffled in the train set, the test set is isolated before any transformations are applied. After training our TerrAI instance, Figure \ref{fig:TSF-TerrAI-TestSet-Results} illustrates the final result on a randomly selected image from the test set of the ITC dataset, whereas the complete results, in terms of three estimation error metrics (namely, RMSE, MAPE, and SMAPE), are listed in the respective rows of Table \ref{tab:terrai_size_vs_energy_vs_accuracy} (row titled ``Baseline''). 
        
        It can be observed that our model has acquired a ``very good representation'' of the 2nd phase of the fertilization process, with an average MAPE of $5.31\%$ per patch, and $9.72\%$ per reconstructed prescription map. Looking at the value of RMSE, we have a score of $12.31$. This means that within an 80$\text{m}^2$ area (i.e., our image patch), our estimations diverge, in average, by 12 $\text{kg} \cdot \text{N} \cdot \text{ha}^{-1}$ from their corresponding ground truth. Compared to the actual prescription map (c.f., Figure \ref{fig:TSF-TerrAI-TestSet-Results}), we observe that our estimations have ``smoother'' transitions towards nearby regions, a behaviour that is partially attributed to the loss function.

    \subsection{%
        \texorpdfstring{
            A Note on the Energy Consumption and CO\textsubscript{2} Emission Savings of TerrAI
        }{
            A Note on the Energy Consumption and CO2 Emission Savings of TerrAI
        }
    }
        By using off-the-shelf energy profiling tools\footnote{For the purposes of this study, we used the GREEN.DAT.AI Benchmark tool by INESC-TEC (patent pending), \url{https://enerframe-greendatai.haslab-dataspace.pt/\#/auth/login}.}, we are able to assess the energy efficiency of TerrAI and identify opportunities for improving energy consumption without compromising model performance. By analysing the profiling results, we discovered that a significant share of the power draw stemmed from the width of the model architecture. Consequently, we revised the model architecture by reducing the amount of trainable parameters. This was achieved by reducing the number of convolutional filters in the encoder decoder (down sampling/up sampling) stages. These changes lower the overall computational load, cut energy consumption, and preserve (or even slightly improve) model generalization.

        For evaluating the overall impact of the aforementioned optimizations, we compare the energy consumption of the ``baseline'' TerrAI (c.f., Section \ref{subsec:Methodology}) against a ``large'' and a ``small'' variant. The difference between these three configurations lies within the number of convolutional channels, which are set to 4 / 8 / 16, 24 / 36 / 48, and 72 / 84 / 96 for the ``small'', ``baseline'', and ``large'' variant, respectively.
        
        The ``Energy consumption'' column of Table \ref{tab:terrai_size_vs_energy_vs_accuracy} illustrates the total energy consumption of TerrAI, across four different runs. It can clearly be observed that reducing the number of channels in the hidden downsampling layers improved its energy efficiency by 49.90\% relative to the baseline. Regarding the trade-off between performance and energy consumption, it can be observed that reducing the number of channels in the hidden downsampling layers increased both RMSE and MAPE metrics by 18.11\% and 15.25\% per patch, respectively. This behaviour can be attributed to the learning capacity of the model, as decreasing the number of channels slightly impacts model performance on the historical data, while further increasing them impacts the generalization capability of the model over non-observed samples. Overall, the ``small'' configuration effectively balances the performance vs. energy consumption trade-off, achieving similar fidelity, albeit with a more energy-efficient model.

        The energy savings of TerrAI can also be expressed in terms of annual CO\(_2\)-equivalent savings that result from the reduced energy demand during model training \cite{PSOMOPOULOS2010485}. Assuming that the model is re-trained periodically (e.g., once per year), the average required energy (in kWh) per run for the three configurations is equals to $E_{\text{run, kWh}}^{\text{small}} = 4.62 \times 10^{-3}$ kWh, $E_{\text{run, kWh}}^{\text{baseline}} = 9.21 \times 10^{-3}$ kWh, and $E_{\text{run, kWh}}^{\text{large}} = 14.66 \times 10^{-3}$ kWh, for the ``small'', ``baseline'', and ``large'' configuration of TerrAI, respectively (simply dividing by 3,600,000 to convert J to kWh). From these values, we obtain the per-run energy savings relative to the baseline model:
        
        \begin{equation}
            \Delta E_{\text{run}}^{\text{baseline, algo}} = \lvert E_{\text{run, kWh}}^{\text{baseline}} - E_{\text{run}, J}^{\text{algo}} \rvert, 
        \end{equation}

        \noindent
        or $\Delta E_{\text{run}}^{\text{baseline, small}} = 4.60 \times 10^{-3}$ kWh, and $\Delta E_{\text{run}}^{\text{baseline, large}} = 5.44 \times 10^{-3}$ kWh, respectively. Since the dataset samples are located in Slovenia and assuming an annual re-training of the model, we use the ``Greenhouse gas emission intensity of electricity generation in Europe'' index\footnote{Greenhouse gas emission intensity of electricity generation in Europe, \url{https://www.eea.europa.eu/en/analysis/indicators/greenhouse-gas-emission-intensity-of-1}. Last visited: 15 December 2025} for the closest available year, or:

        \begin{equation}
            EF_{\text[grid, 2023]}^{\text{avg}} = 0.166\ kg\ CO_2 e / kWh
        \end{equation}

        \noindent
        Applying this factor over the aforementioned energy savings, the annual avoided greenhouse gas emissions are:

        \begin{equation}
            CO_{\text{2, 2023}}^{\text{baseline, algo}} = \Delta E_{\text{run}}^{\text{baseline, algo}} \times EF_{\text{grid, 2023}}
        \end{equation}

        \noindent 
        or, $CO_{\text{2, 2023}}^{\text{baseline, small}} = 0.76$ $g\ CO_2 e$, and $CO_{\text{2, 2023}}^{\text{baseline, large}} = 0.90$ $g\ CO_2 e$ for the ``small'' and ``baseline'' configuration of TerrAI, respectively (see the ``Annual energy savings'' column of Table \ref{tab:terrai_size_vs_energy_vs_accuracy}). It can be observed that the former cuts greenhouse-gas emissions by $15.54$\%, thereby reinforcing the role of TerrAI as an environmentally responsible AI solution for precision agriculture.

\section{Conclusion}\label{sec:conclusion}  
    In this paper, we proposed TerrAI, an efficient CNN-based solution for Targeted Spraying and Fertilization (TSF) over wheat fields. Through an experimental study on a real-world remote-sensing dataset, we demonstrated its efficacy on real-life decision-making processes by evaluating its error across several performance metrics. We also highlighted its ``green'' dimension by measuring energy consumption and the corresponding carbon-emission metrics over three model variants with respect to the number of learnable parameters. 
    
    In the near future, we aim to address the ``blurriness'' in the prescription maps by experimenting with alternative loss functions and further fine-tuning the model architecture and/or feature engineering process, in order to improve the accuracy of TerrAI. Moreover, to generalise across all three fertilisation phases, we aim to exploit on more advanced ML-based techniques, e.g., few-shot learning \cite{PORTO2023100307}, that specialize in learning from few samples. In a parallel line of research, we aim to extend TerrAI to the Federated Learning paradigm \cite{DEMBANI2025110048} and assess the trade-off between privacy, prediction quality, and communication costs.

\section*{Acknowledgment}\addcontentsline{toc}{section}{Acknowledgment}
    \noindent
    This work was supported in part by the Horizon Europe Research and Innovation Programme of the European Union under grant agreement No. 101070416 (Green.Dat.AI; \url{https://greendatai.eu}). The authors would also like to acknowledge INESC-TEC for providing access to their Energy benchmarking tool.

\bibliographystyle{IEEEtran.bst}
\bibliography{bibliography.bib}

\end{document}